\documentclass{article}
\usepackage{iclr2027_conference,times}
\usepackage{amsmath,amssymb,booktabs,tabularx,array}
\usepackage{graphicx}
\usepackage{tikz}
\usepackage{pgfplots}
\pgfplotsset{compat=1.18}
\usetikzlibrary{arrows.meta,positioning,calc}
\usepackage{algorithm}
\usepackage{algpseudocode}
\usepackage{placeins}
\usepackage{hyperref}
\usepackage{url}
\hypersetup{hidelinks,pdftitle={MetaSkill: Reusable Problem-Solving Methods for Adaptive Skill Orchestration in LLM Agents},pdfauthor={Tianxing Wang, Mingming Zhao, Shuai Huang, Chaoyue Niu, Huiyang Xu, Fan Wu, Guihai Chen}}

\newcommand{\method}{\textsc{MetaSkill}}

\newcolumntype{Y}{>{\raggedright\arraybackslash}X}

\title{MetaSkill: Reusable Problem-Solving\\Methods for Adaptive Skill Orchestration\\in LLM Agents}
\author{%
  \textbf{Tianxing Wang\textsuperscript{1}, Mingming Zhao\textsuperscript{2}, Shuai Huang\textsuperscript{2}, Chaoyue Niu\textsuperscript{3},}\\
  \textbf{Huiyang Xu\textsuperscript{4}, Fan Wu\textsuperscript{1}, Guihai Chen\textsuperscript{1}}\\[3pt]
  \small\normalfont\textsuperscript{1}Shanghai Jiao Tong University\\
  \small\normalfont\textsuperscript{2}Huawei Technologies Ltd.\\
  \small\normalfont\textsuperscript{3}Shanghai Jiaotong University, Department of Computer Science\\
  \small\normalfont\textsuperscript{4}China Mobile Communications Company Limited Research Institute%
}
\iclrfinalcopy

\begin{document}
\raggedbottom
\setlength{\floatsep}{7pt plus 2pt minus 1pt}
\setlength{\textfloatsep}{9pt plus 2pt minus 2pt}
\setlength{\intextsep}{7pt plus 2pt minus 1pt}
\maketitle
\lhead{Preprint}
\begin{abstract}
Reusable skills provide specialized procedures, but large language model (LLM) agents must still
determine how to organize and adapt their use.
We introduce \textsc{MetaSkill}, a two-layer framework for reusing
problem-solving methods across tasks and domains.
The method layer selects from 26 curated methods and binds
abstract roles to task objects.
The execution layer uses the instantiated method to identify intermediate
objectives, coordinate native skills, and guide repair or method
reselection.
On 89 SkillsBench tasks and 995 ToolHop examples, MetaSkill outperforms
the evaluated baselines in all six configurations, with accuracy gains
of 3.51--13.48\% over the strongest baseline.
Ablations support explicit method-guided orchestration beyond direct
method injection and further gains from method reselection.
\end{abstract}

\section{Introduction}
\label{sec:intro}

General-purpose LLM agents need to reuse problem-solving knowledge across task types and domains, even when inputs, tools, and deliverables differ substantially. Reusable skills provide specialized procedures and workflows \citep{wang2023voyager,wang2025awm,li2026agentskillos}, and the principles that organize their use offer a further opportunity for reuse. Tasks requiring different operations can still share ways of establishing prerequisites, interpreting intermediate results, and deciding when an approach should change. Reusing these principles gives an agent a coherent method for approaching a new task while selecting the skills appropriate to its domain. We therefore study problem-solving methods that remain useful as task objects and concrete implementations change.

\begin{figure}[t]
\centering
\includegraphics[width=\linewidth]{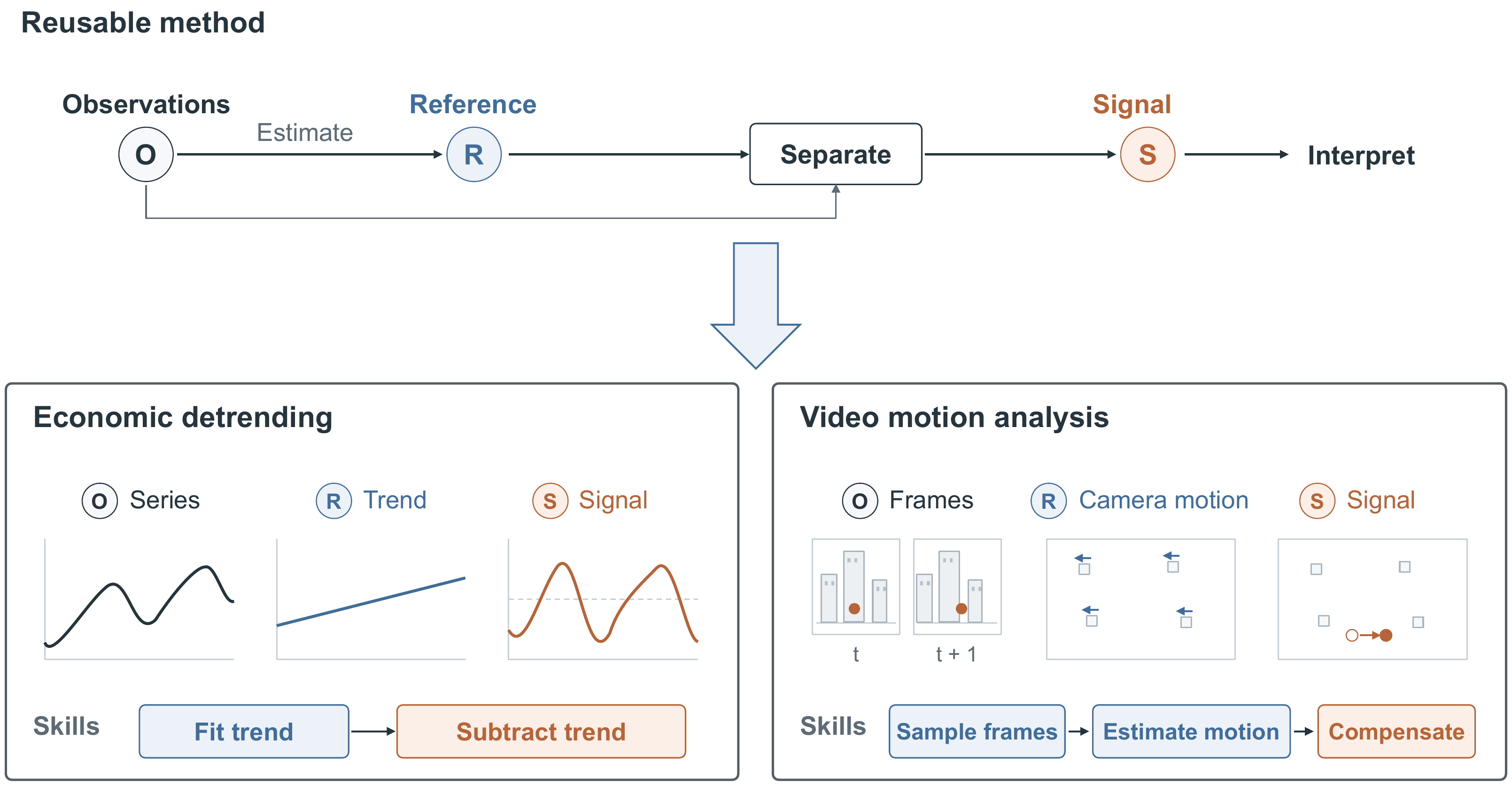}
\caption{Reference-based separation across two domains. Observations ($O$) and an estimated reference ($R$) yield a target-relevant signal ($S$). Economic detrending uses trend fitting and subtraction; video motion analysis uses frame sampling, camera-motion estimation, and compensation. The method supplies the shared relationship, while native skills realize the domain-specific operations.}
\label{fig:method-transfer}
\end{figure}

Consider economic detrending and video motion analysis (Figure~\ref{fig:method-transfer}). The former estimates a long-term trend before analyzing short-term variation; the latter estimates camera motion before identifying independent object motion. Observations ($O$), an estimated reference ($R$), and a target-relevant signal ($S$) play corresponding roles in both cases. Reusing the shared method guides the agent to establish a reliable reference before interpreting the remaining signal, check that separation preserves the variation of interest, and revisit dependent conclusions when the reference changes. This provides a transferable basis for determining which intermediate results to obtain, coordinating the skills that produce them, and avoiding conclusions based on unreliable or outdated results.

Existing work supports reuse at several levels: programs and workflows retain concrete operations and their combinations \citep{wang2023voyager,wang2025awm}, skill orchestration organizes capabilities and their execution dependencies \citep{li2026agentskillos}, and reasoning modules, thought templates, and dynamically selected methodologies provide more abstract guidance \citep{zhou2024selfdiscover,yang2024bot,liu2025com}. These forms of reuse have distinct limitations. Reusing a concrete workflow depends on its original operations and capabilities, limiting direct transfer when a new domain requires different skills. Higher-level reasoning methods can transfer strategies, but the connection from these strategies to native skill execution must still be established for each task: which objects instantiate the method, which intermediate results are prerequisites, and how available skills should produce and check them. This gap limits the influence of reusable methods on actual skill execution. \textsc{MetaSkill} explicitly retains these role relationships, dependencies, and decision principles in reusable method definitions. Binding these definitions to the current task gives skill selection, composition, and checking a shared basis, allowing the same problem-solving logic to guide execution across tasks and domains.

We introduce \textsc{MetaSkill}, a two-layer framework for applying reusable problem-solving methods to skill orchestration across tasks and domains. We construct its library from public task instructions and accompanying skill procedures through LLM-assisted analysis and human curation. We first organize solution processes into method families according to the main work they address, then derive individual methods from shared applicability conditions, dependencies, and decision principles. At runtime, the \emph{method layer} selects a suitable method and binds its abstract roles to task objects. The \emph{execution layer} uses the instantiated method to determine the next required results and organize native skills to obtain them. Feedback guides local repair and supports method reselection when the current approach no longer fits.

Experiments on SkillsBench \citep{li2026skillsbench} and ToolHop \citep{ye2025toolhop} show that \textsc{MetaSkill} achieves the highest accuracy among the evaluated methods in all six configurations. In the SkillsBench ablation, the full framework achieves a task accuracy of 65.17\%, compared with 53.93\% when generating a method separately for each task at runtime and 55.06\% when supplying the same method definitions without explicit method-guided orchestration. These comparisons show that curated, reusable methods provide more effective problem-solving guidance, and that applying this guidance to skill selection, organization, and checking makes better use of that knowledge. The results demonstrate the benefit of reusing problem-solving methods to organize concrete capabilities across heterogeneous tasks.

\section{Related Work}
\label{sec:related}

\subsection{Reusable Problem-Solving Knowledge}

Reusable problem-solving knowledge aims to transfer effective strategies across problems. SELF-DISCOVER selects, adapts, and composes atomic reasoning modules into a task-level reasoning structure that guides different instances of that task~\citep{zhou2024selfdiscover}. Buffer of Thoughts distills high-level thought templates from problem-solving experience and retrieves and instantiates them for new problems~\citep{yang2024bot}. Chain of Methodologies dynamically selects human-defined methodologies to guide reasoning, incorporating execution feedback into subsequent iterations~\citep{liu2025com}.

These studies primarily use reusable knowledge to improve reasoning. \textsc{MetaSkill} focuses on how shared problem-solving methods organize capability use in general-purpose agents across task types and domains. This extends method reuse to capability organization during task execution, allowing the same problem-solving approach to guide tasks supported by different skill sets.

\subsection{Agent Skills and Skill Orchestration}

Agent skills package procedures, knowledge, and workflows as reusable capabilities, while skill orchestration coordinates their use for task completion. AgentSkillOS combines skill discovery with dependency-aware organization to construct task-specific multi-skill workflows~\citep{li2026agentskillos}. SkillPyramid turns existing experience into capabilities for new tasks through hierarchical skill consolidation and composition, supporting the continued evolution of the skill library~\citep{xiong2026skillpyramid}. The original EffiSkill formulation (v1) targets code efficiency optimization and uses Meta Skills to guide the selection and composition of optimization operators, illustrating the role of high-level strategies in domain-specific capability use~\citep{wang2026effiskill}.

These studies advance skill management, composition, and expansion, including through high-level strategy guidance. \textsc{MetaSkill} organizes capability use around problem-solving methods shared across tasks, using the chosen approach to guide skill selection and organization. This provides a reusable methodological basis for orchestration across domains: different skill combinations can implement the same problem-solving approach, and their organization can be revised when that approach changes.

\subsection{Feedback-Driven Agent Adaptation}

Feedback-driven adaptation enables agents to revise decisions based on execution outcomes. ExpeL extracts reusable insights from task experience~\citep{zhao2024expel}, while Reflexion uses verbal reflections to improve subsequent attempts~\citep{shinn2023reflexion}. For adaptation during execution, AdaPlanner revises plans in response to environmental feedback~\citep{sun2023adaplanner}, and TROVE uses runtime traces to validate and edit skill routes while retaining valid continuations~\citep{wang2026trove}.

These studies use feedback to accumulate reusable experience, revise plans, or adjust execution routes. \textsc{MetaSkill} focuses on adaptation at the level of the problem-solving method, examining whether difficulties encountered during execution call for reconsidering the current approach. By connecting method reuse with runtime adaptation, it supports reassessing the selected method as a task progresses and guides subsequent capability use accordingly.
\section{Problem Formulation}
\label{sec:problem}
\label{sec:problem-formulation}

\paragraph{Tasks and capabilities.}
A task instance is
\begin{equation}
    \tau=(x,\mathcal{S}_{\tau},E,B),
    \label{eq:task-setting}
\end{equation}
where $x$ specifies instructions, inputs, constraints, and deliverables; $\mathcal{S}_{\tau}$ is the available native skill library; $E$ supplies tools and an interactive environment; and $B$ limits execution resources. The agent must interpret the requirements, obtain intermediate results, and produce a satisfying deliverable within this budget.

A native skill provides a capability description, procedural guidance, and optional resources. It may cover a complete workflow, including planning and validation, and can support multiple tasks. The agent interprets this guidance through reasoning and tools; a skill need not expose a single callable function.

\paragraph{Reuse across tasks.}
Specialized procedures change with the task, but their reasoning principles can recur. Vehicle rebalancing and transaction analysis involve different entities and operations, yet both require translating rules into explicit conditions and checking conclusions against them. Transferring this strategy requires identifying the relevant state and constraints in the new task, then selecting capabilities to construct and validate the representation.

Decision principles must transfer with these roles. For example, fixing every vehicle's initial load to zero narrows a rebalancing task that permits a bounded initial load. Further solver effort would still optimize the restricted model. A reusable method should therefore guide problem interpretation and checking, as well as the organization of operations.

\paragraph{Objective.}
We seek a shared method library $\mathcal{M}$ whose role relationships and decision principles can guide the use of $\mathcal{S}_{\tau}$ across tasks. Instantiation maps these abstractions to concrete entities and intermediate objectives. Execution observations then determine whether to continue, repair the current work, or reconsider its interpretation. Section~\ref{sec:method} defines the representation and execution process; our experiments compare it with agents already equipped with native skills.

\section{Design and Methodology}
\label{sec:metaskill-methodology}
\label{sec:metaskill-overview}
\label{sec:method}

\begin{figure}[!htbp]
  \centering
  \includegraphics[width=\linewidth]{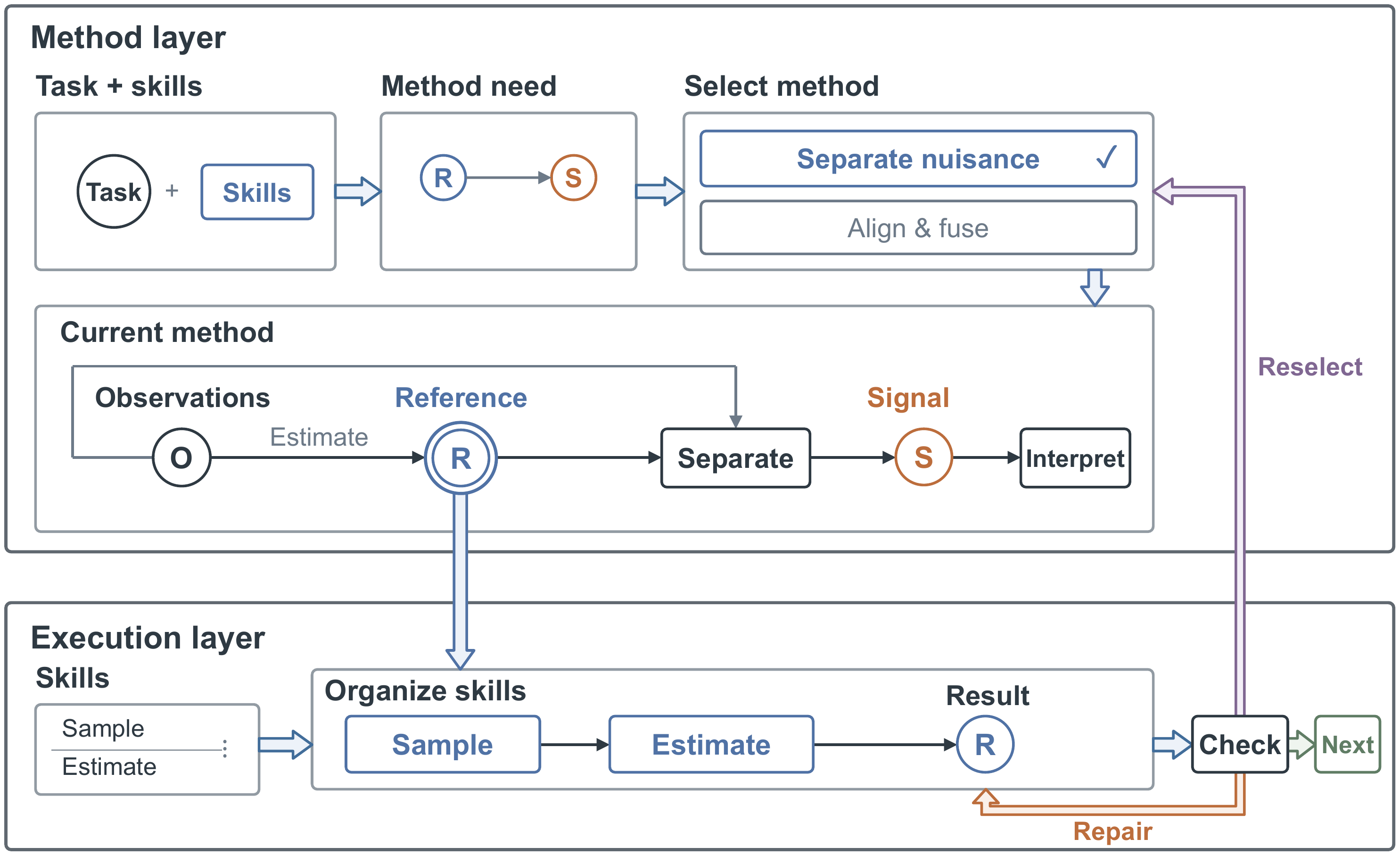}
  \caption{MetaSkill's method and execution layers. Task requirements and available skills inform method selection and role binding. The highlighted reference ($R$) is the current required result, obtained by organizing sampling and estimation skills. A successful check enables progress; feedback can trigger local repair or method reselection.}
  \label{fig:metaskill-layers}
\end{figure}

\method{} connects reusable problem-solving methods with available skills through two layers (Figure~\ref{fig:metaskill-layers}). The method layer identifies the needed guidance, selects a method, and binds its roles to the task. The execution layer obtains the required results by organizing concrete skills. Outcome checks support progress, local repair, or method reselection. Library construction occurs offline, while instantiation and adaptation occur during task execution.

\subsection{Method Layer: Abstraction and Task Instantiation}
\label{sec:metaskill-method-layer}

\subsubsection{Method Library Construction and Representation}
\label{sec:manual-method-extraction}

We construct the library offline from public task descriptions and their accompanying native skills. We first use the task goals and the solutions described in the skills to organize solution processes into seven families according to the main work they require. For example, uncertainty and evidence covers testing competing explanations, representation and transformation covers separating a signal from nuisance variation, and construction and change covers repairs that must preserve surrounding structure. Within each family, we compare when the solutions apply, which intermediate results depend on others, and what guides the choice of the next operation. From these comparisons we extract recurring decision principles. We replace task-specific objects with roles that an agent can bind at runtime and retain the relationships that guide progress, checking, and revision. Native skills supply the concrete procedures.

An LLM assists with source analysis and drafts candidate methods. Human curators compare the candidates across tasks, check their proposed roles and rules against the source descriptions and skills, and revise and select the final entries. This process yields 26 structured methods across the seven families. Appendix~\ref{app:catalog} lists the methods and locates their full definitions and the selection prompt in the supplement.

For example, economic detrending removes long-term trends before analyzing short-term variation, while video analysis compensates for camera motion before identifying independently moving objects. Figure~\ref{fig:method-transfer} expresses their shared relationship as observations ($O$), reference ($R$), and signal ($S$). The corresponding library method, \emph{separate--nuisance--estimate}, establishes the interfering component before interpreting the remaining signal. Its role slots identify the observations, nuisance reference, target, and residual estimator. Filters, alignment algorithms, and thresholds remain with the native skills.

We encode these relationships as role slots, state variables, and policy rules. The documented constraints and recovery conditions inform applicability conditions, invariants, review checks, repair rules, and termination criteria. In the example, checks examine both removed and remaining variation; changing the nuisance estimate reopens dependent results. Table~\ref{tab:compact-method-definition} illustrates the resulting definition.

\begin{table}[htbp]
  \centering
  \caption{Extracted definition of \emph{separate--nuisance--estimate}, abbreviated. $\checkmark$ marks fields included in its discovery card.}
  \label{tab:compact-method-definition}
  \small
  \setlength{\tabcolsep}{5pt}
  \renewcommand{\arraystretch}{1.13}
  \begin{tabularx}{\linewidth}{@{}>{\raggedright\arraybackslash}p{0.18\linewidth}>{\raggedright\arraybackslash}X>{\centering\arraybackslash}p{0.06\linewidth}@{}}
    \toprule
    \textbf{Field} & \textbf{Example content} & \textbf{Card} \\
    \midrule
    Identifier & \texttt{separate-nuisance-estimate} & $\checkmark$ \\
    Description & Account for nuisance effects before estimating the target from residuals. & $\checkmark$ \\
    Entry conditions & Mixed target and nuisance effects; a defensible nuisance estimate. & $\checkmark$ \\
    Exclusions & Do not remove variation merely because it is inconvenient. & $\checkmark$ \\
    \midrule
    Role slots & Observations, target, nuisance reference, residual estimator. & \\
    State variables & Estimate and version, valid region, residuals, uncertainty, dependent results. & \\
    Policy rules & Establish the reference; inspect removed and remaining variation. & \\
    Invariants & Preserve the target; unexplained residuals alone do not identify their cause. & \\
    Review checks & Check sensitivity to the reference and possible removal of target variation. & \\
    Repair rules & Revise the estimate or valid region; recompute affected results. & \\
    Termination & Deliver supported estimates, or explain why separation remains unresolved. & \\
    \bottomrule
  \end{tabularx}
\end{table}

The discovery card contains the identifier, description, entry conditions, and exclusions for candidate comparison. The full definition supplies application rules, stored in \texttt{method.json} and rendered as \texttt{SKILL.md}, alongside source and validation metadata. Appendix~\ref{app:method-schema} describes the schema.

\subsubsection{Selecting and Instantiating a Method for a Task}
\label{sec:metaskill-application}

\paragraph{Identify the needed guidance.}
The agent examines task goals, constraints, and relevant native procedures to identify dependencies, shared results, and revision points. It considers what these relationships mean for the goal: an estimate may remove interference in one task and support prediction in another. Method selection therefore depends on the purpose of the operations, as well as their ordering.

\paragraph{Compare candidates.}
The agent compares discovery cards, then reads relevant definitions to check applicability and useful decision guidance. In video analysis, \emph{separate--nuisance--estimate} addresses the dependency between camera-motion estimation and residual interpretation; alignment and fusion would instead suit combining multiple observations. When native procedures already provide sufficient guidance, the agent can proceed without an added method.

\paragraph{Bind roles to task objects.}
The selected method is instantiated through a concrete role mapping. For video analysis, observations ($O$) become frames and the reference ($R$) becomes estimated camera motion. The target is independently moving objects, whose motion supplies the signal ($S$) after compensation; the residual estimator computes compensated-frame differences. Revising the motion estimate reopens dependent judgments. The agent records the method, selection rationale, role mapping, and scope: either the whole task or a subproblem.

\subsection{Capability Execution: Realization and Feedback}
\label{sec:metaskill-execution-layer}

\subsubsection{Selecting Skills and Organizing Their Execution}
\label{sec:metaskill-orchestration}
\label{sec:metaskill-decisions}

Execution identifies the next required result from task constraints, current evidence, and method rules. The highlighted $R$ in Figure~\ref{fig:metaskill-layers}, for example, requires obtaining a camera-motion estimate before compensation. We call each such unit of work a method ``node.'' Nodes arise as execution progresses and can use multiple skills, invoke a complete native workflow, or reuse existing results. Without an active method, task requirements and native procedures guide the same process.

\paragraph{Select capabilities for the required result.}
The agent checks candidate skills' capabilities, inputs, outputs, and conditions of use. Summaries support discovery; full definitions specify procedures and restrictions. Establishing camera motion, for instance, can require a sampling-and-indexing skill followed by an egomotion-estimation skill. These skills are complementary, whereas two estimators may offer alternative implementations. The agent chooses among alternatives according to available inputs and task constraints.

\paragraph{Coordinate dependencies and shared results.}
The agent orders operations, specifies exchanged outputs, and maintains consistency across them. Sampling indices must correspond to the frames used for motion estimation, compensation, and mask generation. Missing inputs or incompatible outputs trigger revised skill selection or ordering. A skill can support several nodes, and a node can combine several skills; their relationship is determined by the required results.

\paragraph{Check the outcome.}
Before execution, the agent records the expected result, selection rationale, and conditions for changing the plan. It then checks the outcome against task requirements and method rules. Motion parameters, for example, must support valid compensation and residual analysis. Sufficient results enable progress; failures initiate the feedback process below.

\subsubsection{Feedback-Based Repair and Method Reselection}
\label{sec:metaskill-execution-feedback}
\label{sec:metaskill-adaptation}

\paragraph{Localize and repair.}
The agent diagnoses the scope of a failed check. Missing inputs, incorrect arguments, and unsuitable implementations call for execution changes; misinterpreted objects require revised role bindings. In video analysis, an unreliable motion estimate can motivate better sampling or a different estimator. Valid intermediate results are retained, while affected dependencies are checked again or recomputed.

\paragraph{Reconsider the method.}
Method-level revision is triggered when entry conditions cannot be met, an exclusion applies, or the method's checks and repairs cannot support the required judgment. The agent records the conflicting assumption, attempted repairs, and missing evidence, then compares alternatives against the unresolved problem. If residuals remain compatible with both compensation error and true motion, \emph{contrast--test--eliminate} may guide a discriminating test. Selecting it requires feasible observations with different predicted outcomes and an explanation of how they would change the next action. The agent can also return to native procedures when these address the gap.

\paragraph{Resume with updated guidance.}
A method change updates role bindings and checks before execution resumes. Existing frames and sampling indices may remain valid, while judgments based on unreliable compensation require reassessment. The agent records which results are retained or reopened. Execution ends when task and applicable method checks pass, or when the budget or feasible work is exhausted. Appendix~\ref{app:implementation} gives the execution algorithm.

\FloatBarrier
\begingroup\raggedbottom\setlength{\intextsep}{7pt plus 1pt minus 1pt}
\section{Experiments}
\label{sec:experiments}
\subsection{Experimental Setup}
\label{sec:experimental_setup}

\paragraph{Benchmarks and baselines.}
We evaluate on 89 SkillsBench tasks (87 main and 2 supplementary) after excluding
10 tasks with setup failures~\citep{li2026skillsbench}, and on 995 multi-hop
tool-use examples from ToolHop~\citep{ye2025toolhop}. We compare MetaSkill with
WithSkill, BoT~\citep{yang2024bot}, CoM~\citep{liu2025com}, and
AgentSkillOS~\citep{li2026agentskillos}. WithSkill is the skills-only baseline. The LLM receives the native skills available for a task (tool descriptions on ToolHop) and uses its own reasoning to choose and sequence actions. We reimplement CoM for each benchmark. SkillOpt~\citep{rao2026skillopt} uses verifier source and feedback to compile a skill offline for repeated runs of the same task. Our comparison evaluates how agents organize the available skills during task execution.

\paragraph{Configuration.}
Both benchmarks use DeepSeek-V4.1-Flash (\texttt{deepseek-v4-flash} via
\url{https://api.deepseek.com}) and GPT-4.1-mini
(\texttt{gpt-4.1-mini-2025-04-14} via \url{https://api.chatanywhere.tech}).
On SkillsBench, we pair each model with Anthropic's
\href{https://github.com/anthropics/claude-code}{Claude Code} and the
\href{https://github.com/badlogic/pi-mono}{Pi coding agent} (PI Agent).
Runs use the frozen per-task time and resource budgets with bounded token use,
checking, and repair. ToolHop allows nine execution steps per example.
A run that exhausts this budget without a valid final-answer file is incomplete.

\paragraph{Metrics.}
We report task accuracy for the SkillsBench comparison and ablations, and strict
accuracy on ToolHop. Both main tables report mean tokens per example in thousands (k)
and absolute accuracy differences (\%), computed by subtracting the two accuracy percentages.

\subsection{Main Results}
\label{sec:main_results}
\label{sec:toolhop_results}
MetaSkill achieves the highest accuracy in all six configurations. Its margin over
the strongest baseline ranges from 4.50\% to 13.48\% on SkillsBench and from
3.51\% to 4.72\% on ToolHop (Tables~\ref{tab:main_results} and~\ref{tab:toolhop_results}).
Method guidance thus improves performance both on tasks with native skills and
on multi-hop tool use.

\begin{table}[!htbp]
\begingroup
\centering
\caption{Accuracy (\%) and average token usage per task (k; 1k = 1,000 tokens) on the 89-task SkillsBench subset. $\Delta$ is the accuracy of MetaSkill minus that of the corresponding method. Best accuracy within each configuration is shown in bold.}
\label{tab:main_results}
\vspace{5pt}
\small
\renewcommand{\arraystretch}{1.15}
\setlength{\tabcolsep}{3pt}
\begin{tabular*}{\linewidth}{@{\extracolsep{\fill}}lrrrrrr@{}}
\toprule
& \multicolumn{3}{c}{\textbf{GPT-4.1-mini}} & \multicolumn{3}{c}{\textbf{DeepSeek-V4.1-Flash}} \\
\cmidrule(lr){2-4}\cmidrule(lr){5-7}
\textbf{Method} & \textbf{Acc. (\%)} & $\boldsymbol{\Delta}$ \textbf{(\%)} & \textbf{Tokens (k)} & \textbf{Acc. (\%)} & $\boldsymbol{\Delta}$ \textbf{(\%)} & \textbf{Tokens (k)} \\
\midrule
\multicolumn{7}{@{}l}{\textit{Claude Code}} \\
\addlinespace[3pt]
WithSkill & 6.74 & 10.11 & 648.03 & 53.93 & 14.61 & 1438.50 \\
CoM~\citep{liu2025com} & 8.99 & 7.86 & 1190.37 & 48.31 & 20.23 & 1487.43 \\
BoT~\citep{yang2024bot} & 11.24 & 5.61 & 923.11 & 50.56 & 17.98 & 2624.09 \\
AgentSkillOS~\citep{li2026agentskillos} & 10.11 & 6.74 & 1020.01 & 55.06 & 13.48 & 3208.24 \\
\textbf{MetaSkill} & \textbf{16.85} & --- & 811.41 & \textbf{68.54} & --- & 4255.59 \\
\midrule
\multicolumn{7}{@{}l}{\textit{PI Agent}} \\
\addlinespace[3pt]
WithSkill & 10.11 & 7.87 & 754.76 & 52.81 & 12.36 & 1420.46 \\
CoM~\citep{liu2025com} & 11.24 & 6.74 & 1201.91 & 47.19 & 17.98 & 1072.33 \\
BoT~\citep{yang2024bot} & 12.36 & 5.62 & 880.15 & 60.67 & 4.50 & 1323.48 \\
AgentSkillOS~\citep{li2026agentskillos} & 8.99 & 8.99 & 830.75 & 51.69 & 13.48 & 2327.29 \\
\textbf{MetaSkill} & \textbf{17.98} & --- & 857.46 & \textbf{65.17} & --- & 2487.53 \\
\bottomrule
\end{tabular*}
\endgroup

\end{table}
\begin{table}[!htbp]
\begingroup
\centering
\caption{Strict accuracy (\%) and average token usage per example (k; 1k = 1,000 tokens) on 995 ToolHop examples. $\Delta$ is the accuracy of MetaSkill minus that of the corresponding method. Best accuracy for each model is shown in bold.}
\label{tab:toolhop_results}
\vspace{5pt}
\small
\renewcommand{\arraystretch}{1.15}
\setlength{\tabcolsep}{3pt}
\begin{tabular*}{\linewidth}{@{\extracolsep{\fill}}lrrrrrr@{}}
\toprule
& \multicolumn{3}{c}{\textbf{GPT-4.1-mini}} & \multicolumn{3}{c}{\textbf{DeepSeek-V4.1-Flash}} \\
\cmidrule(lr){2-4}\cmidrule(lr){5-7}
\textbf{Method} & \textbf{Acc. (\%)} & $\boldsymbol{\Delta}$ \textbf{(\%)} & \textbf{Tokens (k)} & \textbf{Acc. (\%)} & $\boldsymbol{\Delta}$ \textbf{(\%)} & \textbf{Tokens (k)} \\
\midrule
WithSkill & 40.70 & 11.26 & 5.10 & 42.51 & 22.21 & 11.55 \\
CoM~\citep{liu2025com} & 43.72 & 8.24 & 19.09 & 48.04 & 16.68 & 23.74 \\
BoT~\citep{yang2024bot} & 47.24 & 4.72 & 18.87 & 61.21 & 3.51 & 33.27 \\
AgentSkillOS~\citep{li2026agentskillos} & 41.61 & 10.35 & 37.45 & 57.19 & 7.53 & 77.18 \\
\addlinespace[3pt]
\textbf{MetaSkill} & \textbf{51.96} & --- & 21.48 & \textbf{64.72} & --- & 38.43 \\
\bottomrule
\end{tabular*}
\endgroup

\end{table}

\paragraph{Token cost by model and agent.}
Figure~\ref{fig:accuracy-cost} plots accuracy against mean tokens per example.
On SkillsBench, MetaSkill uses the most tokens with DeepSeek, particularly when
paired with the more exploratory Claude Code agent. The execution traces show
what additional calls can accomplish. In the date-parsing case, inspecting the
intermediate output identifies a missing month contribution, so the agent can
correct the date format and rerun the failed step (Appendix~\ref{app:case-studies}).
With GPT-4.1-mini, MetaSkill instead outperforms BoT with fewer tokens under
both agents. Under Claude Code, for example, accuracy rises by 5.61\% while
mean token use falls by 12.1\%. Guiding the executor toward a specific
intermediate result can therefore improve accuracy even when it spends less.

\begin{figure}[!htbp]
\centering
\includegraphics[width=.90\linewidth]{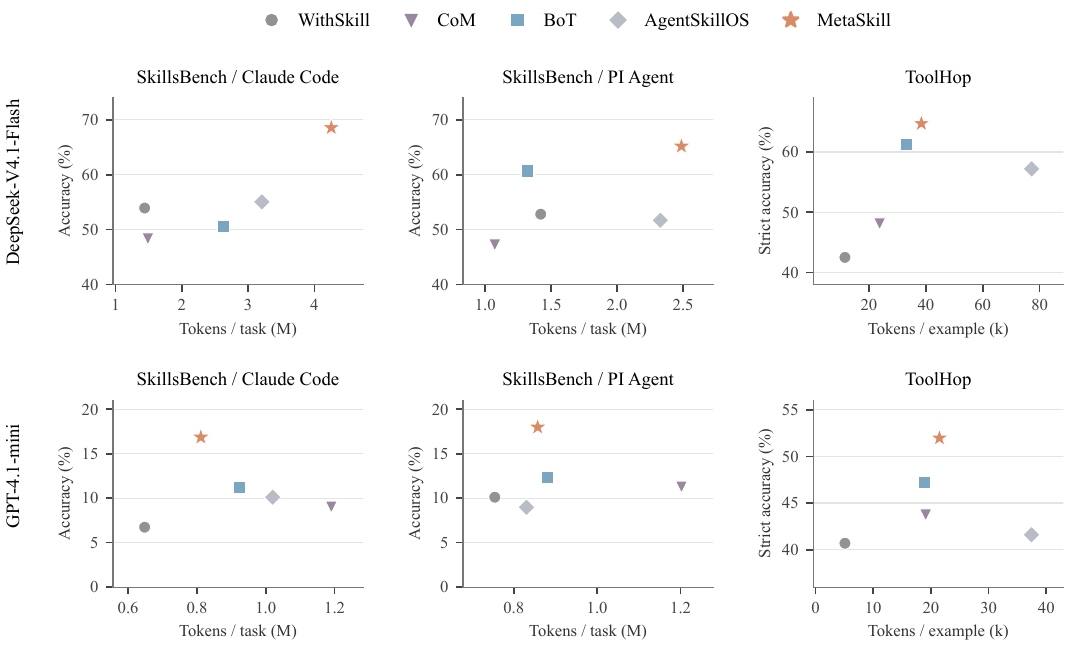}
\caption{Accuracy and average token consumption across the configurations in Tables~\ref{tab:main_results} and~\ref{tab:toolhop_results}. Each point represents one method; higher accuracy and lower token consumption are preferable. Axes use configuration-specific ranges, with token usage in millions (M) for SkillsBench and thousands (k) for ToolHop.}
\label{fig:accuracy-cost}
\end{figure}

\FloatBarrier
\subsection{Ablation Study}
\label{sec:ablation}

\paragraph{Configurations.}
We compare seven configurations on 89 SkillsBench tasks using PI Agent with DeepSeek (Figure~\ref{fig:ablation}). The no-skills condition removes both native skills and method guidance. The skills-only condition restores the native skills, and direct injection additionally supplies method definitions as text. Generic orchestration retains planning, checking, and repair while removing the method library and its rules. Online generation replaces the library with a method generated for each task under the same budget. The no-reselection condition retains the initial method and local repair but disables method switching. The full system combines the offline library with runtime method guidance and reselection.

\paragraph{Offline method library.}
Replacing the offline library with a method generated for each task under the same budget lowers accuracy from 65.17\% to 53.93\%. Removing the library and its method-specific guidance while retaining generic planning, checking, and repair also lowers accuracy (57.30\%). These results support the value of offline construction. A library of methods abstracted across tasks gives the agent more useful guidance than it can generate for a single task during execution.

\begin{figure}[H]
\centering
\includegraphics[width=\linewidth]{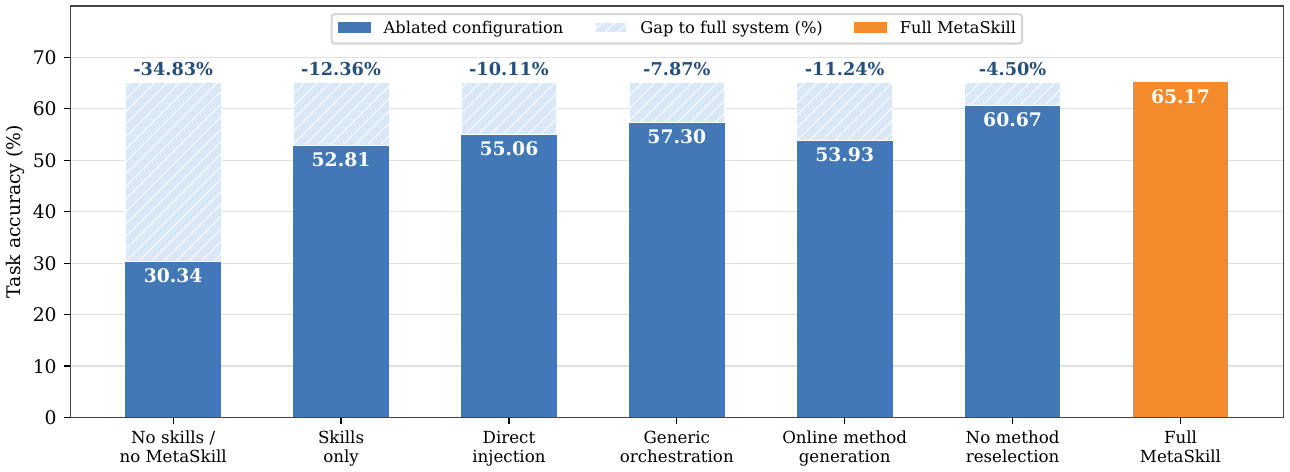}
\caption{Ablation on 89 SkillsBench tasks with PI Agent and DeepSeek-V4.1-Flash. Solid bars show task accuracy; hatched extensions show the gap to full MetaSkill, with labels giving the absolute accuracy difference (\%).}
\label{fig:ablation}
\end{figure}

\paragraph{Runtime method layer.}
With the library available, directly injecting its text achieves 55.06\% accuracy, only slightly above native skills alone (52.81\%). Applying the methods through the full runtime layer raises accuracy by 10.11\% over direct injection. Within that layer, disabling method reselection costs 4.50\% even when local repair remains available. Thus the runtime contribution has two parts. The agent must use the selected method to guide execution and be able to revise that choice as it gathers evidence.

\FloatBarrier
\subsection{Results Analysis}
\label{sec:fine-grained-analysis}
\label{sec:error-case-analysis}

Table~\ref{tab:toolhop-breakdown} compares CoM, BoT, AgentSkillOS, and MetaSkill on the same 995 ToolHop examples using DeepSeek-V4.1-Flash. It groups tasks first by annotated sub-task count (3, 4, or $\geq 5$) and then by domain, pooling domains with fewer than 30 examples as Other. Figure~\ref{fig:toolhop-error-modes} breaks down strict errors for the same four methods. We record an \emph{answer transfer failure} when the last tool result contains the reference answer but the agent submits a final response that fails strict scoring. The other categories are incomplete runs and completed runs in which both the submission and the last-tool fallback fail.

\begin{table}[H]
\centering
\caption{ToolHop strict accuracy (\%) with DeepSeek-V4.1-Flash by sub-task count and domain (995 examples). Domains with fewer than 30 examples are grouped as Other.}
\label{tab:toolhop-breakdown}
\vspace{5pt}
\small
\renewcommand{\arraystretch}{1.08}
\setlength{\tabcolsep}{5pt}
\begin{tabular*}{\linewidth}{@{\extracolsep{\fill}}lrrrr@{}}
\toprule
\textbf{Group} & \textbf{CoM} & \textbf{BoT} & \textbf{AgentSkillOS} & \textbf{MetaSkill} \\
\midrule
\multicolumn{5}{@{}l}{\textit{Sub-tasks}} \\
\addlinespace[2pt]
3 & 49.30 & \textbf{60.51} & 54.91 & 60.05 \\
4 & 53.54 & 64.59 & 63.74 & \textbf{67.14} \\
$\geq 5$ & 36.45 & 57.01 & 50.93 & \textbf{70.09} \\
\midrule
\multicolumn{5}{@{}l}{\textit{Domain}} \\
\addlinespace[2pt]
Computing & 28.57 & 28.57 & 26.53 & \textbf{57.14} \\
Film & 61.75 & 75.41 & 72.68 & \textbf{81.42} \\
Genealogy & 24.64 & 34.78 & 27.54 & \textbf{36.96} \\
History & 25.34 & \textbf{44.52} & 41.78 & 41.78 \\
Linguistics & 51.35 & 62.16 & 45.95 & \textbf{70.27} \\
Literature & 73.33 & 75.56 & \textbf{84.44} & 82.22 \\
Mathematics & 66.07 & 71.43 & 71.43 & \textbf{73.21} \\
Other & 53.87 & 70.97 & 65.81 & \textbf{71.29} \\
Publishing & 77.42 & 87.10 & 80.65 & \textbf{96.77} \\
\bottomrule
\end{tabular*}
\end{table}

MetaSkill's advantage is concentrated in longer chains. MetaSkill solves 70.09\% of examples with at least five sub-tasks, ahead of the strongest baseline in that group (BoT, 57.01\%). This group accounts for 28 of the 35 additional correct answers over BoT. Under the common nine-step budget, MetaSkill delivers a final answer on every example, while CoM leaves 418 unfinished and BoT leaves 69. Final verification remains a challenge. In the counting case, the agent resolves the right entity but applies the wrong counting rule (Appendix~\ref{app:case-failure}). Checking what the final operation computes remains important even when all dependencies have been resolved.
\begin{figure}[H]
\centering
\includegraphics[width=\linewidth]{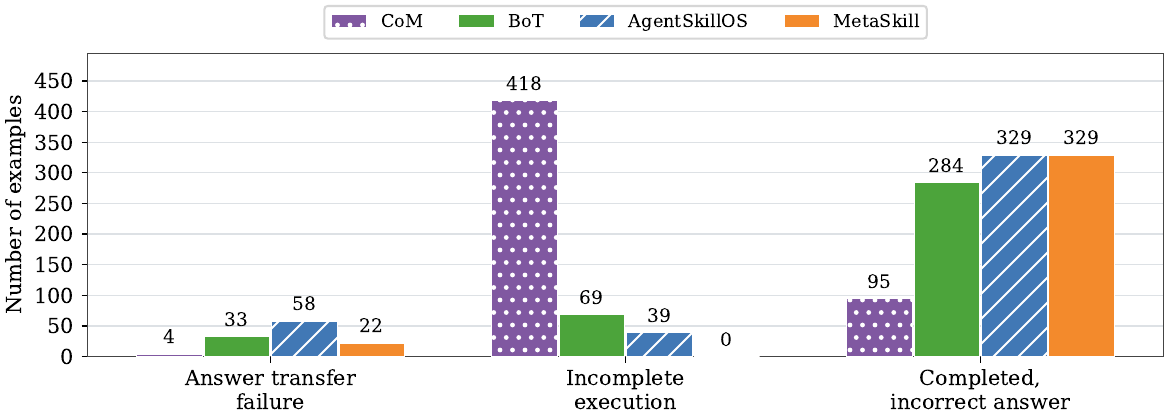}
\caption{ToolHop error outcomes with DeepSeek-V4.1-Flash. An answer transfer failure occurs when the last tool result contains the reference answer but the final submission fails strict scoring. Total strict errors are 517 for CoM, 386 for BoT, 426 for AgentSkillOS, and 351 for MetaSkill.}
\label{fig:toolhop-error-modes}
\end{figure}

\endgroup

\section{Conclusion}
\label{sec:conclusion}
MetaSkill enables agents to reuse problem-solving methods across tasks and coordinate native skills to obtain the intermediate results each method requires. It binds a method's roles to the current task, uses the resulting checks to guide execution, and revises the method when feedback challenges the initial choice. On SkillsBench, the ablations distinguish these functions. Supplying a method as text yields only a small gain over native skills, and a generic check-and-repair loop still falls short of method-specific execution. Allowing the agent to reconsider its initial method adds a further benefit beyond local repair.

The same methods transfer to ToolHop, where most of the gain over BoT comes from longer dependency chains under a fixed step limit. The computational trade-off depends on the executor. With DeepSeek and Claude Code, diagnosing and retrying failed operations can add substantial work. With GPT-4.1-mini, MetaSkill achieves higher SkillsBench accuracy than BoT with fewer tokens under both agent frameworks. The remaining ToolHop failures suggest a specific direction for improving the method library. Alongside checks that each prerequisite has been obtained, methods need checks that the final operation answers the question asked. Execution traces that distinguish these failures could guide the refinement of those checks.

\FloatBarrier

\section*{AI Use Statement}
We used generative AI tools to assist with drafting and polishing the
manuscript, including improving language, clarity, and organization,
and preparing illustrative figures.
The authors reviewed and revised the AI-assisted text and figures
and take full responsibility for the final manuscript.

\section*{Reproducibility Statement}
Section~\ref{sec:method} specifies the method representation and execution algorithm. Section~\ref{sec:experimental_setup} describes the benchmarks, model and agent configurations, and metrics. Appendices~\ref{app:method-schema} and~\ref{app:implementation} detail the schema and runtime semantics, and Appendix~\ref{app:case-studies} provides trace-based cases.

\bibliography{references}

@misc{wang2023voyager,
  title = {{Voyager}: An Open-Ended Embodied Agent with Large Language Models},
  author = {Wang, Guanzhi and Xie, Yuqi and Jiang, Yunfan and Mandlekar, Ajay and Xiao, Chaowei and Zhu, Yuke and Fan, Linxi and Anandkumar, Anima},
  year = {2023},
  howpublished = {arXiv preprint arXiv:2305.16291},
  url = {https://arxiv.org/abs/2305.16291}
}

@misc{li2026agentskillos,
  title = {Organizing, Orchestrating, and Benchmarking Agent Skills at Ecosystem Scale},
  author = {Li, Hao and Mu, Chunjiang and Chen, Jianhao and Ren, Siyue and Cui, Zhiyao and Zhang, Yiqun and Bai, Lei and Hu, Shuyue},
  year = {2026},
  howpublished = {arXiv preprint arXiv:2603.02176},
  url = {https://arxiv.org/abs/2603.02176}
}

@inproceedings{zhou2024selfdiscover,
  title = {{Self-Discover}: Large Language Models Self-Compose Reasoning Structures},
  author = {Zhou, Pei and Pujara, Jay and Ren, Xiang and Chen, Xinyun and Cheng, Heng-Tze and Le, Quoc V. and Chi, Ed H. and Zhou, Denny and Mishra, Swaroop and Zheng, Huaixiu Steven},
  year = {2024},
  booktitle = {Advances in Neural Information Processing Systems},
  volume = {37},
  url = {https://proceedings.neurips.cc/paper_files/paper/2024/hash/e41efb03e20ca3c231940a3c6917ef6f-Abstract-Conference.html},
  doi = {10.52202/079017-4004}
}

@inproceedings{yang2024bot,
  title = {Buffer of Thoughts: Thought-Augmented Reasoning with Large Language Models},
  author = {Yang, Ling and Yu, Zhaochen and Zhang, Tianjun and Cao, Shiyi and Xu, Minkai and Zhang, Wentao and Gonzalez, Joseph E. and Cui, Bin},
  booktitle = {Advances in Neural Information Processing Systems},
  volume = {37},
  year = {2024},
  url = {https://proceedings.neurips.cc/paper_files/paper/2024/hash/cde328b7bf6358f5ebb91fe9c539745e-Abstract-Conference.html}
}

@inproceedings{zhao2024expel,
  title = {{ExpeL}: {LLM} Agents Are Experiential Learners},
  author = {Zhao, Andrew and Huang, Daniel and Xu, Quentin and Lin, Matthieu and Liu, Yong-Jin and Huang, Gao},
  booktitle = {Proceedings of the AAAI Conference on Artificial Intelligence},
  volume = {38},
  pages = {19632--19642},
  year = {2024},
  doi = {10.1609/aaai.v38i17.29936},
  url = {https://ojs.aaai.org/index.php/AAAI/article/view/29936}
}

@inproceedings{wang2025awm,
  title = {Agent Workflow Memory},
  author = {Wang, Zora Zhiruo and Mao, Jiayuan and Fried, Daniel and Neubig, Graham},
  year = {2025},
  booktitle = {Proceedings of the 42nd International Conference on Machine Learning},
  volume = {267},
  series = {Proceedings of Machine Learning Research},
  url = {https://proceedings.mlr.press/v267/wang25bx.html},
  pages = {63897--63911},
  publisher = {PMLR}
}

@inproceedings{shinn2023reflexion,
  title = {{Reflexion}: Language Agents with Verbal Reinforcement Learning},
  author = {Shinn, Noah and Cassano, Federico and Gopinath, Ashwin and Narasimhan, Karthik and Yao, Shunyu},
  booktitle = {Advances in Neural Information Processing Systems},
  volume = {36},
  year = {2023},
  url = {https://proceedings.neurips.cc/paper_files/paper/2023/hash/1b44b878bb782e6954cd888628510e90-Abstract-Conference.html},
  doi = {10.52202/075280-0377}
}

@inproceedings{sun2023adaplanner,
  title = {{AdaPlanner}: Adaptive Planning from Feedback with Language Models},
  author = {Sun, Haotian and Zhuang, Yuchen and Kong, Lingkai and Dai, Bo and Zhang, Chao},
  booktitle = {Advances in Neural Information Processing Systems},
  volume = {36},
  year = {2023},
  url = {https://proceedings.neurips.cc/paper_files/paper/2023/hash/b5c8c1c117618267944b2617add0a766-Abstract.html},
  doi = {10.52202/075280-2537}
}

@misc{wang2026trove,
  title = {{TROVE}: Adaptive Agent Skill Orchestration via Trace-Grounded Route Validation and Editing},
  author = {Wang, Tianxing and Zhao, Mingming and Huang, Shuai and Xu, Huiyang and Niu, Chaoyue and Liu, Shengzhong and Wu, Fan},
  year = {2026},
  howpublished = {arXiv preprint arXiv:2609.05019},
  url = {https://arxiv.org/abs/2609.05019}
}

@inproceedings{liu2025com,
  title = {Chain of Methodologies: Scaling Test Time Computation without Training},
  author = {Liu, Cong and Wu, Jie and Wu, Weigang and Chen, Xu and Lin, Liang and Zheng, Wei-Shi},
  booktitle = {Findings of the Association for Computational Linguistics: ACL 2025},
  year = {2025},
  pages = {5298--5312},
  publisher = {Association for Computational Linguistics},
  doi = {10.18653/v1/2025.findings-acl.276},
  url = {https://aclanthology.org/2025.findings-acl.276/}
}

@misc{xiong2026skillpyramid,
  title = {{SkillPyramid}: A Hierarchical Skill Consolidation Framework for Self-Evolving Agents},
  author = {Xiong, Yuan and Miao, Ziqi and Chen, Qian and Li, Lijun and Deng, Ken and Wang, Yequan and He, Shizhu and Zhao, Jun and Liu, Kang},
  year = {2026},
  howpublished = {arXiv preprint arXiv:2606.03692},
  url = {https://arxiv.org/abs/2606.03692}
}

@misc{wang2026effiskill,
  title = {{EffiSkill}: Agent Skill Based Automated Code Efficiency Optimization},
  author = {Wang, Zimu and Shi, Yuling and Li, Mengfan and Liu, Zijun and Zhang, Jie M. and Wan, Chengcheng and Gu, Xiaodong},
  year = {2026},
  howpublished = {arXiv preprint arXiv:2603.27850v1},
  url = {https://arxiv.org/abs/2603.27850v1}
}

@misc{li2026skillsbench,
  title = {{SkillsBench}: Benchmarking How Well Agent Skills Work Across Diverse Tasks},
  author = {Li, Xiangyi and Liu, Yimin and Chen, Wenbo and others},
  year = {2026},
  howpublished = {arXiv preprint arXiv:2602.12670},
  url = {https://arxiv.org/abs/2602.12670}
}

@inproceedings{ye2025toolhop,
  title = {{ToolHop}: A Query-Driven Benchmark for Evaluating Large Language Models in Multi-Hop Tool Use},
  author = {Ye, Junjie and Du, Zhengyin and Yao, Xuesong and Lin, Weijian and Xu, Yufei and Chen, Zehui and Wang, Zaiyuan and Zhu, Sining and Xi, Zhiheng and Yuan, Siyu and Gui, Tao and Zhang, Qi and Huang, Xuanjing and Chen, Jiecao},
  booktitle = {Proceedings of the 63rd Annual Meeting of the Association for Computational Linguistics (Volume 1: Long Papers)},
  year = {2025},
  pages = {2995--3021},
  publisher = {Association for Computational Linguistics},
  doi = {10.18653/v1/2025.acl-long.150},
  url = {https://aclanthology.org/2025.acl-long.150/}
}

@misc{rao2026skillopt,
  title = {{SkillOpt}: Trajectory-Derived, Verifier-Grounded Compilation of {LLM}-Agent Skills},
  author = {Rao, Rohan Gurunandan and Srinivas, Vineeth Kalluru},
  year = {2026},
  howpublished = {OpenReview manuscript},
  url = {https://openreview.net/pdf/8a1c72f069408db6661f7b1709de339257449694.pdf}
}
\bibliographystyle{iclr2027_conference}
\clearpage
\appendix
\section{Method Schema and Representative Entries}
\label{app:method-schema}

Table~\ref{tab:method-schema-mapping} maps the semantic components in Section~\ref{sec:manual-method-extraction} to the detailed fields used in the manuscript's method schema. Identity, provenance, and validation metadata accompany the fields below. Role relationships and dependencies are expressed in natural-language role descriptions and policies.

\begin{table}[htbp]
  \centering
  \caption{Core method components and their detailed schema fields.}
  \label{tab:method-schema-mapping}
  \small
  \setlength{\tabcolsep}{5pt}
  \renewcommand{\arraystretch}{1.18}
  \begin{tabularx}{\linewidth}{@{}>{\raggedright\arraybackslash}p{0.20\linewidth}>{\raggedright\arraybackslash}p{0.32\linewidth}>{\raggedright\arraybackslash}X@{}}
    \toprule
    \textbf{Core component} & \textbf{Schema fields} & \textbf{Function} \\
    \midrule
    Applicability & Entry conditions; exclusions & Specify when the method is relevant and which conclusions or uses are outside its scope. \\
    \addlinespace[4pt]
    Problem abstraction & Role slots; state variables & Identify entities and relationships to bind, and information relevant to the next decision. \\
    \addlinespace[4pt]
    Decision guidance & Policy rules; invariants; review checks; repair rules; termination & Guide the next required result, preserve reasoning constraints, and govern checking, revision, and completion. \\
    \bottomrule
  \end{tabularx}
\end{table}

\subsection{Method Families and Catalog}
\label{app:catalog}
Table~\ref{tab:catalog} lists the library used for method selection. Each family groups methods by the work a solution process requires. In the accompanying archive \path{method_library.zip}, each \path{methods/<method-id>/method.json} contains the complete definition for its table entry. The planning branch of \texttt{build\_context\_v4} in \path{prompts/context_v4.py} constructs the selection prompt from the task and available method cards.
\begin{table}[H]
\centering\small
\caption{The seven method families and their 26 MetaSkills.}
\label{tab:catalog}
\begin{tabularx}{\linewidth}{@{}p{.26\linewidth}X@{}}
\toprule
\textbf{Family} & \textbf{MetaSkills} \\
\midrule
Uncertainty and evidence & probe-update-advance, contrast-test-eliminate, trace-link-resolve, triangulate-resolve-synthesize \\[3pt]
Representation and transformation & parse-map-reconcile, recover-from-redundancy, align-synchronize-fuse, separate-nuisance-estimate, segment-transform-reassemble \\[3pt]
Formalization and search & rule-model-recompute, constrain-propagate-search, coarse-localize-refine, anchor-expand-improve, construct-counterexample-refine, filter-compare-rank \\[3pt]
Construction and change & contract-decompose-integrate, localize-preserve-repair, map-preserve-migrate \\[3pt]
Resource and performance & partition-balance-merge, measure-bottleneck-optimize \\[3pt]
Composition and presentation & claim-counterclaim-restructure, template-bind-instantiate, structure-render-localize-polish \\[3pt]
Dynamics and control & identify-calibrate-generalize, receding-horizon-act-replan, feedback-error-stabilize \\[3pt]
\bottomrule
\end{tabularx}
\end{table}
\subsection{Representative Methods}

\paragraph{Recover from redundancy.}
The roles are unknowns, redundant observations, relationships, and tolerances. The method identifies independent relationships, determines which unknowns are identifiable, and solves them in dependency order. Remaining relationships provide cross-checks against source precision and units. Review examines uniqueness and conflicting constraints; a contradiction directs attention to the source relationship before a recovered value is revised. The method completes when identifiable required values have been recovered and checked, and remaining ambiguity or contradiction is described explicitly.

\paragraph{Contrast--test--eliminate.}
The roles are explanations, predictions, discriminating checks, and observations. The state retains live explanations, predicted outcomes, uncertainty, and unresolved alternatives. A useful check has outcomes that separate explanations under stated assumptions. An observation eliminates only explanations contradicted within the tested scope; insufficient sensitivity or compatible predictions leave alternatives unresolved. Review examines confounding, equivalent predictions, and dependencies of reused evidence.

\paragraph{Parse--map--reconcile.}
The roles are source fields, canonical entities or units, mapping rules, and a conflict policy. The state retains parsed records, mapping support, ambiguous fields, and a reconciliation ledger. Parsing syntax is separated from assigning meaning. Canonical values retain their source and transformation history, and an uncertain mapping remains uncertain rather than being silently filled. A changed mapping requires recomputation of affected downstream records.

\paragraph{Rule--model--recompute.}
The roles bind public rules to decision state, model predicates, and output checks. The method separates task constants from permitted decisions and temporary strategy restrictions. It then uses an appropriate modeling capability and recomputes the serialized result against the rules. Feasibility, incumbent quality, and proven optimality are reviewed separately. A feasible witness tests whether the formulation admits a particular allowed case; domain coverage requires checking the correspondence between the public rules and model constraints.

\paragraph{Contract--decompose--integrate.}
This method organizes component work around interfaces and dependencies. It calls for a minimal real end-to-end interaction before broad implementation and preserves a complete native procedure when suitable. Review exercises the final serialized artifact through its intended consumer, checking constructor, shape, and behavioral requirements in the actual integration path.

\section{Implementation Semantics}
\label{app:implementation}

\begin{algorithm}[H]
\caption{Task execution with an existing MetaSkill library}
\label{alg:metaskill-solving}
\small
\begin{algorithmic}[1]
\Require Task, available skills, method library, environment, and budget
\State Examine requirements and skill dependencies; select a method or none.
\State Record the selection rationale and bind roles when a method is selected.
\While{budget permits and feasible work remains}
 \State Identify the next result using task requirements, current evidence, and method rules.
 \State Select skills by capabilities and prerequisites, or reuse valid results.
 \State Organize dependencies and shared outputs; revise skill choices as needed.
 \State Record the expected result and change conditions; execute and check the outcome.
 \If{final checks pass for task requirements and applicable method rules}
  \State \Return deliverables and execution records
 \ElsIf{the outcome is insufficient}
  \State Diagnose the failure using feedback and earlier repair outcomes.
  \If{the evidence calls for reconsidering the method}
   \State Compare alternatives, including no added method, against the unresolved problem.
   \State Update the method, role bindings, and checks when justified and permitted.
  \Else
   \State Revise inputs, skills, execution, task interpretation, or role bindings as needed.
  \EndIf
  \State Keep valid results; reopen affected conclusions and recompute as needed.
 \EndIf
\EndWhile
\State \Return available results and unresolved conditions
\end{algorithmic}
\end{algorithm}

\paragraph{Selection and state are model-interpreted.}
The planning report selects one method identifier or none and supplies all required role bindings. The LLM interprets their suitability for the task, while the runtime validates identifiers and nonempty bindings. Method state observations can use catalog state names or additional labels; the implementation records matched and unmatched labels separately for trace inspection.

\paragraph{Method context during delegation.}
An active method remains available in subsequent prompts during delegated work, and its review contract still applies. In delegate mode, the agent uses a native procedure to obtain the required result. Selecting no active method retains the full system's available catalog and execution guidance; a method-free experimental condition instead omits method resources from the agent's context.

\paragraph{Evidence tracks applicability to artifacts.}
The runtime records fingerprints of public file artifacts and dependencies used by checks. Supported directory deliverables are fingerprinted from their members and contents. A changed artifact or dependency invalidates affected evidence. Interpretation revision retains the obligation identity and source and reopens satisfaction for review under the revised reading. These records track evidence freshness and execution provenance.

\paragraph{Repair limits are scoped.}
The default permits one method switch and one targeted repair after a failed review or closure. Role rebinding within the same method is accounted for separately from switching methods. Routine tool debugging may occur within a coherent action. All work remains subject to the overall task budget. The final record retains available artifacts and internal closure status separately for external evaluation.

\FloatBarrier
\section{Case Studies}
\label{app:case-studies}

This appendix details the execution cases supporting the analysis in Section~\ref{sec:fine-grained-analysis}. ToolHop examples are identified by their dataset IDs and use the selected DeepSeek-V4.1-Flash trajectories underlying Table~\ref{tab:toolhop_results}. The SkillsBench example is an archived PI Agent execution used to illustrate how a method coordinates multiple skills. We examine the task, observed failure or uncertainty, intervention, and resulting outcome.

\subsection{Diagnosing a Date-Parsing Mismatch}
\label{app:case-date}
ToolHop example 202 asks for the digit sum of a retrieved death date. MetaSkill selects \emph{trace-link-resolve} and records the chain from the named person to the father, the death date, and the numerical transformation. The first call to \texttt{date\_digit\_sum} on ``11 July 1302'' returns 8. Review identifies a conflict with the required computation, and repair requests component-level output: day sum 2, month sum 0, and year sum 6. This isolates the missing month contribution. Repeating the call with ``11/07/1302'' and explicit DMY formatting produces components 2, 7, and 6, yielding the correct answer 15. BoT submits no final answer and AgentSkillOS returns 18. The useful intervention is the choice of a diagnostic observation followed by a targeted input correction, while retaining the already established identity and date.

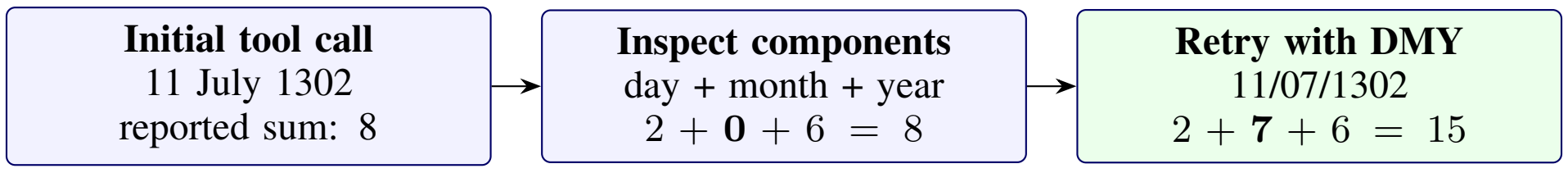
\begin{figure}[H]
\centering
\begin{tikzpicture}[panel/.style={draw=blue!40!black,fill=blue!5,rounded corners=2pt,text width=0.26\linewidth,align=center,font=\small,inner sep=4pt},>=Stealth]
\node[panel] (a) {\textbf{Initial tool call}\\11 July 1302\\reported sum: 8};
\node[panel,right=4mm of a] (b) {\textbf{Inspect components}\\day + month + year\\$2+\mathbf{0}+6=8$};
\node[panel,right=4mm of b,fill=green!8] (c) {\textbf{Retry with DMY}\\11/07/1302\\$2+\mathbf{7}+6=15$};
\draw[->] (a)--(b);\draw[->] (b)--(c);
\end{tikzpicture}
\caption{ToolHop example 202. Inspecting the contributions isolates the missing month. Only the date format changes on the second call.}
\label{fig:case-date}
\end{figure}

\subsection{Repairing a Letter-Counting Parameter}
\label{app:case-count}
In example 151, BoT and AgentSkillOS return 0 repeated letters for the first name of a screenwriter. MetaSkill initially makes the same error. Its selected \emph{trace-link-resolve} method prompts review of the film-credit join, which returns two screenwriters sharing the first name ``David.'' Review reopens the chain because the identity is not uniquely resolved. During repair, the agent rechecks the credits and calls the letter-count tool with \texttt{ignore\_case=true}; the tool returns two occurrences of ``d,'' changing the answer from 0 to 1. Renewed checking reopens the tool call, and the repair stage corrects its case-sensitivity parameter.

\subsection{Checking the Final Answer Representation}
\label{app:case-format}
Example 115 follows a five-step chain from film identification to a character's surname and a binary character code. BoT, AgentSkillOS, and MetaSkill reach the relevant code. BoT and AgentSkillOS submit \texttt{01110000}, whereas the selected MetaSkill run submits \texttt{1110000}, following the experiment's instruction to omit leading zeroes from numeric answers. This formatting change gives MetaSkill strict credit on an example where BoT and AgentSkillOS both fail. End-to-end checking must cover both the evidence chain and the requested answer representation.

\subsection{Preserving Constraints in CAD Repair}
\label{app:case-cad}
The archived computer-aided design (CAD) task \texttt{ada-bathroom-plan-repair} illustrates a different use of the method layer. MetaSkill chooses \emph{localize-preserve-repair}, mapping observed accessibility violations to editable regions and identifying the bathtub, stable element identifiers, and original CAD context as protected content. It coordinates drawing extraction, accessibility checks, and geometric repair. The saved change log records a 7.893-inch toilet displacement, grab-bar adjustments, and repositioning of the turning circle, while retaining the door and bathtub. The run produces both the repaired DXF and synchronized layout and change records, and receives an official score of 1.0; the archived PI Agent AgentSkillOS execution receives 0.0. The trace illustrates how explicit preservation constraints guide several skills toward a consistent artifact.

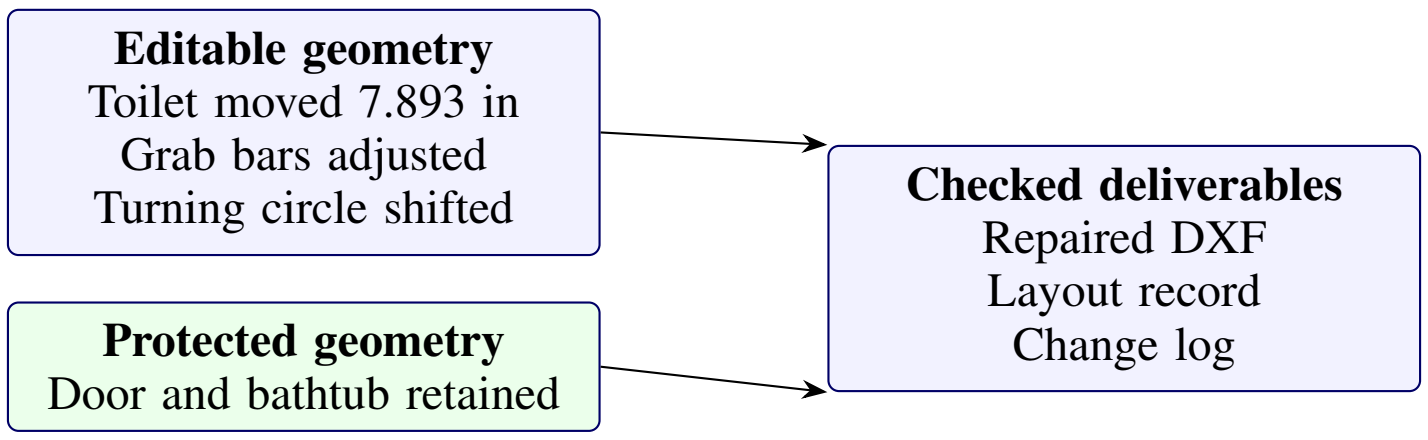
\begin{figure}[H]
\centering
\begin{tikzpicture}[panel/.style={draw=blue!40!black,fill=blue!5,rounded corners=2pt,text width=0.26\linewidth,align=center,font=\small,inner sep=4pt},>=Stealth]
\node[panel] (a) {\textbf{Editable geometry}\\Toilet moved 7.893 in\\Grab bars adjusted\\Turning circle shifted};
\node[panel,below=3mm of a,fill=green!8] (b) {\textbf{Protected geometry}\\Door and bathtub retained};
\node[panel,right=15mm of a,yshift=-9mm] (c) {\textbf{Checked deliverables}\\Repaired DXF\\Layout record\\Change log};
\draw[->] (a.east) -- (c.north west);\draw[->] (b.east) -- (c.south west);
\end{tikzpicture}
\caption{SkillsBench CAD repair. The change log distinguishes edited geometry from protected elements and accompanies the revised drawing.}
\label{fig:case-cad}
\end{figure}

\subsection{A Failure of Semantic Validation}
\label{app:case-failure}
In ToolHop example 35, MetaSkill correctly resolves the film, producer, and surname, then accepts a combination-counting tool's output of 6. Its review passes because the calls form a supported chain, but the submitted answer still fails the benchmark. Here, repeating the successful retrieval path does not test whether the final operation implements the intended counting semantics. The case exposes a remaining error mode: validating evidence provenance without independently validating the meaning of the terminal computation.

\end{document}